\documentclass[10pt,twocolumn,letterpaper,hyphens]{article}
\usepackage{multirow}
\usepackage{graphicx}
\usepackage{booktabs}
\usepackage{colortbl}
\usepackage[accsupp]{axessibility}
\usepackage{pifont}
\usepackage{float}
\usepackage{iccv}              % To produce the CAMERA-READY version
\definecolor{iccvblue}{rgb}{0.21,0.49,0.74}
\usepackage[pagebackref,breaklinks,colorlinks,allcolors=iccvblue]{hyperref}

\newcommand{\StyleGan}[1]{\textcolor{BlueViolet}{#1}}
\newcommand{\Diffuison}[1]{\textcolor{BrickRed}{#1}}
\newcommand{\Gan}[1]{\textcolor{ForestGreen}{#1}}
\def\paperID{1623} % *** Enter the Paper ID here
\def\confName{ICCV}
\def\confYear{2025}

\title{CanonSwap: High-Fidelity and Consistent Video Face Swapping via Canonical Space Modulation}
\makeatletter
\newcommand{\printfnsymbol}[1]{%
  \textsuperscript{\@fnsymbol{#1}}%
}
\renewcommand*{\@fnsymbol}[1]{\ensuremath{\ifcase#1\or *\or \dagger\or \ddagger\or
   \mathsection\or \mathparagraph\or \|\or **\or \dagger\dagger
   \or \ddagger\ddagger \else\@ctrerr\fi}}

\author{
    Xiangyang Luo$^{1,2}$\thanks{Intern at IDEA} \quad Ye Zhu$^{2}\thanks{Corresponding authors}$ \quad Yunfei Liu$^{2}$ \quad Lijian Lin$^{2}$ \quad Cong Wan$^{3}$ \quad Zijian Cai$^{3}$ \\ Shao-Lun Huang$^{1}$\printfnsymbol{2} \quad Yu Li$^{2}$\printfnsymbol{3}  \\
    $^1$Tsinghua Shenzhen International Graduate School, Tsinghua University \\ $^2$International Digital Economy Academy (IDEA) \quad $^3$Xi'an Jiaotong University\\
}

\begin{document}
\twocolumn[{%
\renewcommand\twocolumn[1][]{#1}%
\maketitle
\begin{center}
    \centering
    \captionsetup{type=figure}
    \includegraphics[width=1\textwidth]{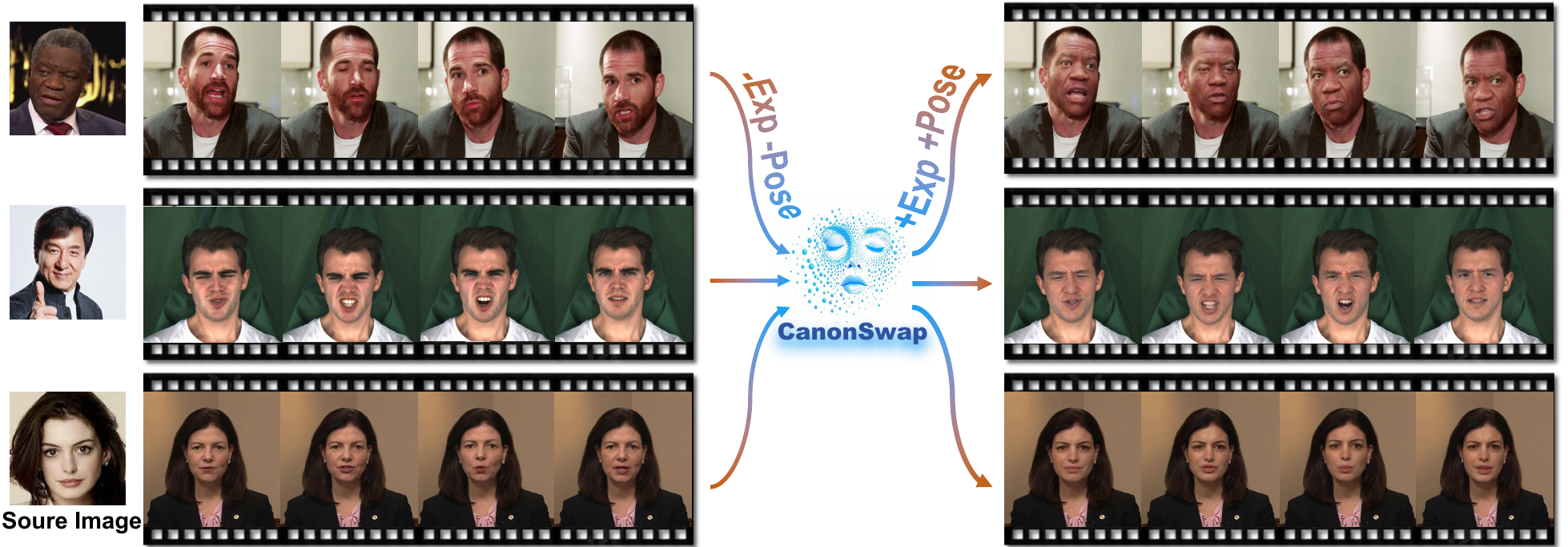}
    \captionof{figure}{CanonSwap decouples motion information from appearance by first transforming the target video into a canonical space for face swapping, then reintroducing its original motion. This process ensures stable, temporally consistent results while accurately preserving motion alignment.
    % preserving the dynamic attributes of the video.
    }
    \label{fig:teaser}
\vspace{-6pt}
\end{center}%
}]
{\let\thefootnote\relax\footnotetext{{$^{*}$ Intern at IDEA. ~ $^{\ddagger}$ Project lead. ~ $^{\dagger}$ Corresponding authors. }}}
\begin{abstract}

Video face swapping aims to address two primary challenges: effectively transferring the source identity to the target video and accurately preserving the dynamic attributes of the target face, such as head poses, facial expressions, lip-sync, \etc.
Existing methods mainly focus on achieving high-quality identity transfer but often fall short in maintaining the dynamic attributes of the target face, leading to inconsistent results. 
We attribute this issue to the inherent coupling of facial appearance and motion in videos. 
To address this, we propose CanonSwap, a novel video face-swapping framework that decouples motion information from appearance information. 
Specifically, CanonSwap first eliminates motion-related information, enabling identity modification within a unified canonical space. Subsequently, the swapped feature is reintegrated into the original video space, ensuring the preservation of the target face's dynamic attributes. 
To further achieve precise identity transfer with minimal artifacts and enhanced realism, we design a Partial Identity Modulation module that adaptively integrates source identity features using a spatial mask to restrict modifications to facial regions.
Additionally, we introduce several fine-grained synchronization metrics to comprehensively evaluate the performance of video face swapping methods.
Extensive experiments demonstrate that our method significantly outperforms existing approaches in terms of visual quality, temporal consistency, and identity preservation. Our project page are publicly available at \url{https://luoxyhappy.github.io/CanonSwap/}. 

\end{abstract}    
\section{Introduction}
\label{sec:intro}

% With the rapid development of digital media technology, video face swapping has gained significant attention in various applications, from entertainment and film production to virtual communication. While image-based face swapping has made considerable progress, video face swapping presents unique challenges, particularly in maintaining temporal consistency and handling pose variations. Traditional face swapping methods \cite{chen2020simswap,li2019faceshifter} often struggle with temporal stability, producing noticeable jittering artifacts when facial poses change throughout video sequences. These methods are inherently sensitive to pose variations, as they attempt to perform face swapping directly in the original pose space.

 % With the rapid development of digital media technology, video face swapping has gained significant attention in various applications, ranging from entertainment and film production to virtual communication. This emerging field faces unique challenges, particularly in maintaining temporal consistency and handling pose variations over time. 

% With the rapid development of digital media technology, video face swapping has gained significant attention in various applications, ranging from entertainment and film production~\cite{film} to privacy preservation\cite{ciftci2023my, mahajan2017swapitup}. 
% Unlike image-based face swapping, video face swapping task faces unique challenges, which not only require replace target face with given source image, but also need smooth transitions, maintaining dynamic facial movements to avoid flickering and jitter.

With the rapid advancement of digital media technology, video face swapping has garnered considerable attention across a wide range of applications, including entertainment~\cite{humanaes}, film production~\cite{film}, and privacy protection~\cite{ciftci2023my, mahajan2017swapitup}. Unlike image-based face swapping, the video face swapping task is more challenging. It not only requires replacing the target face with a given source image but also demands seamless transitions and the preservation of dynamic facial movements to avoid flickering and jitter.

%  With the rapid development of digital media technology, video face swapping has gained significant attention in various applications, ranging from entertainment and film production to virtual communication. Unlike image-based face swapping, video face swapping introduces additional layers of complexity. In videos, not only must the identity transformation be accurate on a per-frame basis, but it must also be consistent across time to avoid flickering and jittering artifacts. Maintaining temporal coherence while handling dynamic variations in pose, expression, and lighting conditions presents a formidable challenge.

% Furthermore, video face swapping requires robust methods that can adapt to subtle changes in motion and preserve the natural flow of facial movements. These challenges emphasize the need for specialized techniques that extend beyond the capabilities of traditional image-based approaches, which often fall short when applied directly to video sequences.

Most existing methods primarily focus on the effectiveness of identity swapping, aiming to achieve high similarity and fidelity between the swapped face and the source image. These approaches can generally be categorized into two main types: GAN-based~\cite{gan} and diffusion-based methods.
% However, most current methods are image-based, and these methods can be categorized into GAN-based~\cite{gan} approaches and diffusion-based~\cite{denosing} approaches. 
GAN-based methods \cite{chen2020simswap,li2019faceshifter, liu2023fine, nirkin2022fsganv2, karras2019style, zhu2024stableswap, codeswap} typically inject identity features into the target image via latent space manipulations or channel-wise normalization, achieving impressive identity transfer. 
Diffusion-based methods~\cite{kim2022diffface,zhao2023diffswap, REFace, faceadapter} reformulate face swapping as a conditional inpainting task, using attribute and keypoint conditions to guide the generation process and refine facial details. 
However, these methods often disrupt the inherent attributes of the face, such as facial expressions and lip movements. Moreover, when applied to videos, they struggle to maintain consistency across frames, resulting in flickering and artifacts.
% Both methods operate directly on the target image, where motion and appearance are tightly coupled, leading to inevitable motion changes after swapping. While such modifications are negligible for still images, they cause temporal inconsistencies and degrade video quality.
% Consequently, when extended to videos, they inevitably alter motion and cause jittering—an issue tolerable in still images but detrimental to temporal consistency in video face swapping.
% However, most current methods are image-based, and these methods can be categorized into GAN-based approaches and diffusion-based approaches. Traditional face swapping approaches \cite{chen2020simswap,li2019faceshifter} have made considerable progress for static images, but when applied frame by frame to videos, they often struggle with temporal stability, producing noticeable jittering artifacts as facial poses change throughout the sequence.
Recent advances in video diffusion models have led to new video face swapping methods~\cite{chen2024hifivfs, wang2025dynamicface, shao2024vividface}. These methods can generate temporal consistent videos. However, these methods often come with substantial computational overhead and may compromise the preservation of original facial dynamics, such as head pose and facial expression. 
% This limitation can lead to degraded synchronization between the swapped face and the target video, particularly in scenarios requiring precise pose alignment.

% To address these challenges, we propose a novel framework that fundamentally decouples pose variations from the face swapping process by operating in a canonical space. Our method first normalizes facial poses into a canonical space, where we introduce a partial identity modulation module that dynamically modifies convolution weights based on source identity features. This module, coupled with a spatial mask for targeted region modification, enables precise face swapping while minimizing artifacts and enhancing realism. The transformed results are then accurately warped back to their original poses, ensuring both temporal stability and precise pose control.

To address the aforementioned challenges, we argue that a stable video face swapping framework requires disentangling facial attributes, specifically separating identity-related information (\eg, appearance) from identity-agnostic information (\eg, facial motion). 
Based on this insight, we propose a novel video face swapping framework, termed CanonSwap. 
Our approach begins by extracting facial motion information from the video, and uses a warping-based method to map the face from the original space into a unified canonical space. 
Subsequently, face swapping modulation is performed in this canonical space. 
Finally, the swapped result is projected back into the original video space to restore the target's inherent facial dynamics, leading to temporal consistency and stability in the generated video, please refer to Fig.~\ref{fig:teaser}.

% To address these challenges, we propose a novel framework that fundamentally disentangles facial motion from appearance during the face swapping process by operating in a canonical space. Our approach begins by warping the input video from its original space into a canonical space where facial poses are disentangled from appearance features. This transformation isolates the static appearance of the face from dynamic motion cues, thereby providing a stable basis for accurate face swapping. Once face swap performed, the transformed features are warped back to their original space to ensure precise alignment and temporal stability.

% To mitigate the influence of non-facial regions on the face swapping process and enhance the swapping quality in the canonical space, we introduce a Partial Identity Modulation module that adaptively incorporates source identity features into the target's appearance. 
% This module employs a spatial mask to constraint the identity modifications exclusively to identity-relative regions, thereby preventing unwanted alterations in non-facial areas. 
% By selectively enhancing local features, our framework achieves high-fidelity identity transfer while minimizing artifacts and preserving fine details. 
% By combining motion-appearance decoupling with targeted identity modulation, our method produces natural-looking face swaps that preserve authentic facial features and ensure smooth, continuous video results.

To mitigate the influence of non-facial regions on the face swapping process and enhance the swapping quality, we introduce a Partial Identity Modulation (PIM) module. This module adaptively integrates source identity features into the target's appearance while employing a spatial mask to constrain identity modifications exclusively to identity-relevant regions. 
PIM prevents unwanted alterations in non-facial areas, thus ensuring that only the necessary facial attributes are modified. 
By adaptively integrating source identity features into the target's appearance, our framework achieves high-fidelity identity transfer, minimizing artifacts and preserving fine-grained details. 

Furthermore, since there exist limited video face swapping evaluation benchmarks, we introduce a comprehensive set of fine-grained evaluation metrics. These include novel synchronization measures, detailed eye-related metrics (such as gaze direction and eye aperture dynamics), and temporal consistency assessments. 

% Combining motion-appearance decoupling with targeted identity modulation, our method produces natural-looking face swaps that maintain authentic facial dynamics and ensure temporally smooth and consistent video results.

% Furthermore, considering the complex nature of video face swapping evaluation, we introduce a comprehensive set of fine-grained metrics specifically designed for video scenarios. These include novel synchronization metrics that assess the alignment between swapped faces and original movements, detailed eye-related measurements covering both gaze direction and eye aperture dynamics, and enhanced temporal consistency metrics. These metrics provide a more thorough evaluation framework for video face swapping methods.

% Furthermore, to address the complexity of evaluating video face swapping, we introduce a suite of fine-grained metrics including novel synchronization measures, detailed eye-related metrics (gaze direction and eye aperture dynamics), and temporal consistency assessments.

Extensive experiments demonstrate that our approach significantly outperforms existing methods. Our contributions can be summarized as follows:
\begin{itemize}
\item We introduce CanonSwap, a canonical space transformation framework that decouples facial motion and facial appearance, achieving both high-quality identity swapping and stable temporal consistency results.
\item We propose a PIM module that achieves accurate identity transfer to the facial region while preserving unwanted regions through partial adaptive weight modification.
\item We introduce comprehensive fine-grained evaluation metrics specifically designed for video face swapping, providing a more detailed assessment of synchronization, eye dynamics, and temporal consistency.
\item Experimental results demonstrate superior performance in terms of visual quality, temporal consistency, and identity preservation compared to existing methods.
\end{itemize}
\section{Related Work}
\label{sec:relate}

\subsection{Image Face Swapping}

Face swapping has attracted significant research interest due to its wide range of practical applications. Early approaches primarily relied on classical image processing techniques and three-dimensional morphable models (3DMMs) \cite{nirkin2018face, bitouk2008face}, which often resulted in visibly artificial swaps. The introduction of generative adversarial networks (GANs) marked a turning point. Early works like FSGAN \cite{nirkin2018face} leveraged GANs for face reenactment and blending, yet struggled with preserving the target's authentic attributes. This limitation spurred the development of AdaIN-based methods \cite{adain, chen2020simswap, li2019faceshifter, gao2021information, Shiohara_2023_ICCV, xu2022high}, which extract identity features from pre-trained face recognition models and fuse them with target features in the latent space.

To further enhance image quality, several works~\cite{mega, xu2022styleswap, liu2023fine} have leveraged StyleGAN~\cite{karras2019style} to boost face swapping performance. Some other approaches~\cite{zhu2024stableswap, codeswap} utilize VQGAN~\cite{esser2021taming} and proposed a multi-stage training method. With the advent of diffusion models, methods such as DiffSwap~\cite{zhao2023diffswap}, DiffFace~\cite{kim2022diffface}, and REFace~\cite{REFace} train diffusion models for face swapping from scratch.
% , yet achieve limited success in terms of visual quality. 
Face Adapater~\cite{faceadapter} introduces an adapter with pretrained diffusion models to achieve high-fidelity face swapping.

However, directly swapping faces inevitably alters the motion due to the coupling between facial appearance and motion. Although such modifications may be negligible in static image face swapping, they may lead to temporal inconsistencies and degraded results in video face swapping.

\subsection{Video Face Swapping}

With the development of video diffusion models~\cite{animatediff, svd}, recent works 
like DynamicFace~\cite{wang2025dynamicface}, VividFace~\cite{shao2024vividface}, and HiFiVFS~\cite{chen2024hifivfs} have attempted to address temporal consistency in video face swapping through temporal attention mechanism~\cite{attention}. However, their experimental results indicate limitations in preserving precise pose and expression dynamics, which are crucial for applications requiring accurate audio-visual synchronization. Additionally, video diffusion-based methods, while prioritizing temporal consistency, often require substantial computational resources and multiple conditioning signals, making them less practical for efficient applications.

In contrast, our method addresses both temporal stability and attribute preservation by operating in a canonical space, effectively decoupling motion from appearance. Despite adopting a frame-by-frame approach, our method achieves robust video face swapping while maintaining precise pose control and temporal consistency within a relatively computationally efficient framework.
% \subsection{Face Warping}
% Face warping refers to the geometric transformation of facial images to different poses or canonical spaces, which plays a crucial role in face-related tasks. Recent advances in portrait animation have demonstrated effective approaches for face warping through various motion representations. Methods like FOMM \cite{siarohin2019first} and Face vid2vid \cite{wang2021facevid2vid} utilize implicit keypoints to capture facial movements and generate optical flow for warping. Face vid2vid particularly extends this concept by introducing 3D implicit keypoints, enabling free-view portrait animation. TPSM \cite{zhao2022thin} further improves warping flexibility through nonlinear thin-plate spline transformation, especially effective for large-scale motions.

% While these methods primarily focus on transferring poses between source and driving images for portrait animation, our approach adapts the warping mechanism for a different purpose. Specifically, we leverage the warping technique from portrait animation to transform face images to and from a canonical space, which serves as a crucial preprocessing step for our face swapping framework. This canonical space transformation effectively normalizes pose variations, enabling more robust video face swapping.

\subsection{Motion Appearance Decoupling}
Motion appearance decoupling refers to the process of separating the dynamic motion information from the static appearance features of facial images, a critical operation in our framework. Recent advances in portrait animation have demonstrated effective techniques for capturing facial motion using keypoints~\cite{landmark1, landmark2, FOMM, facev2v, megaportraits}, semantic segmentation~\cite{nirkin2022fsganv2, freehand}, and 3DMMs~\cite{3d1, 3d2, 3d3,tokenface,liu2025teaser}, as well as generating optical flow for precise warping. Methods like FOMM \cite{FOMM} and Face vid2vid \cite{facev2v} utilize implicit keypoints to model facial movements, while Face vid2vid extends this idea with 3D implicit keypoints to support free-view portrait animation~\cite{anytalk}. Additionally, TPSM \cite{TPSM} employs nonlinear thin-plate spline transformations to handle large-scale motions flexibly. With the development of diffusion models~\cite{ldm}, animation can be reformulated as a conditional inpainting task that integrates both appearance and motion conditions~\cite{animateanyone, spf, disco, humanmotion}.

In contrast to these methods, which primarily aim to transfer poses between source and driving images for animation, our approach repurposes the warping mechanism to decouple motion from appearance. By transforming face images into a canonical space, we effectively decouple pose variations and isolate appearance features from dynamic motion cues. This decoupling serves as a crucial preprocessing step for our face swapping framework, enabling more robust and temporally consistent video face swapping, as well as accurate pose alignment with the target face.
\section{Method}
\label{method}
\begin{figure}
    \centering
    \includegraphics[width=\linewidth]{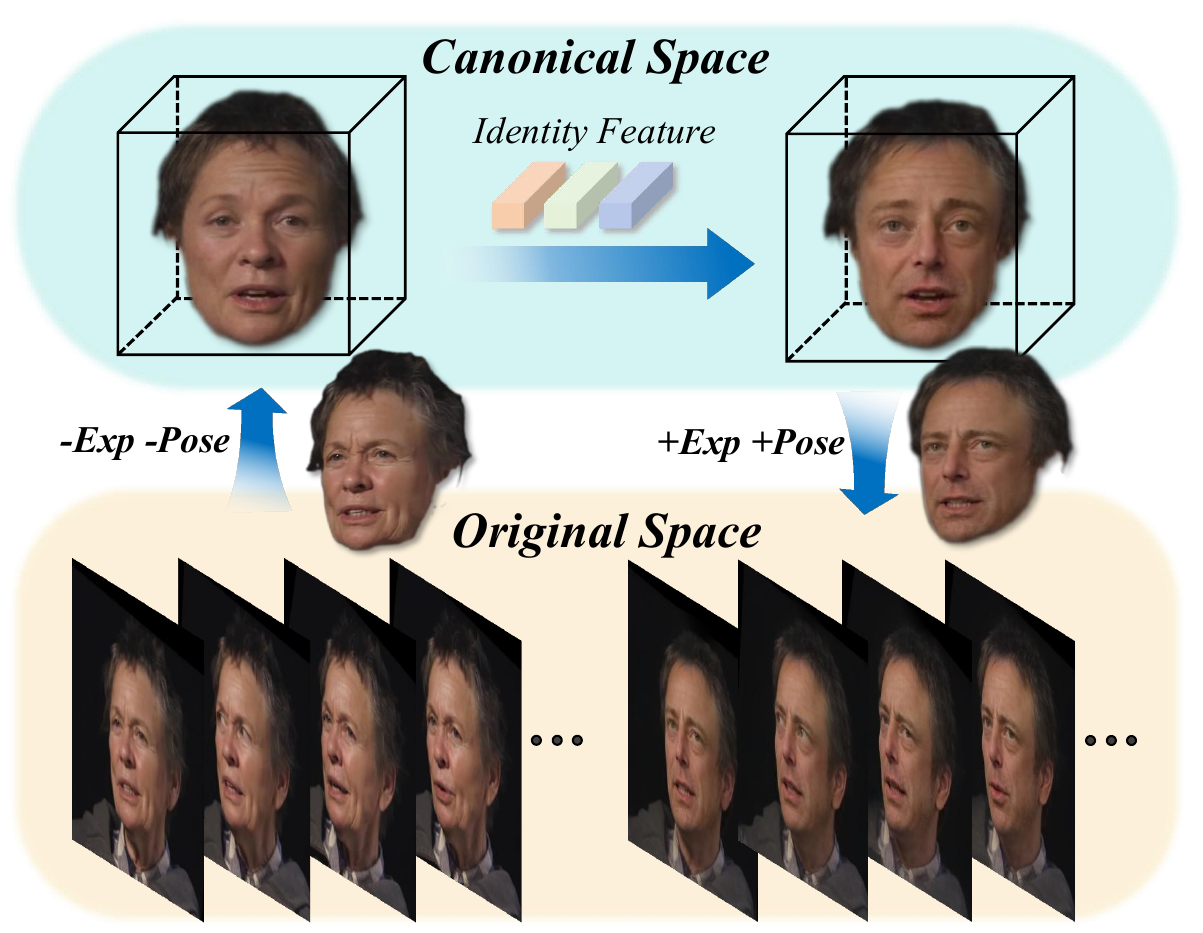}
    \caption{Conceptual illustration of CanonSwap. We transform video frames from the original space to a canonical space to decouple motion information. After performing face swapping in the canonical space, we warp the results back to the original space, achieving precise motion preservation and video consistency.}
    \label{fig:method}
\end{figure}
% \vspace{-1pt} 
\begin{figure*}
    \centering
    \includegraphics[width=\linewidth]{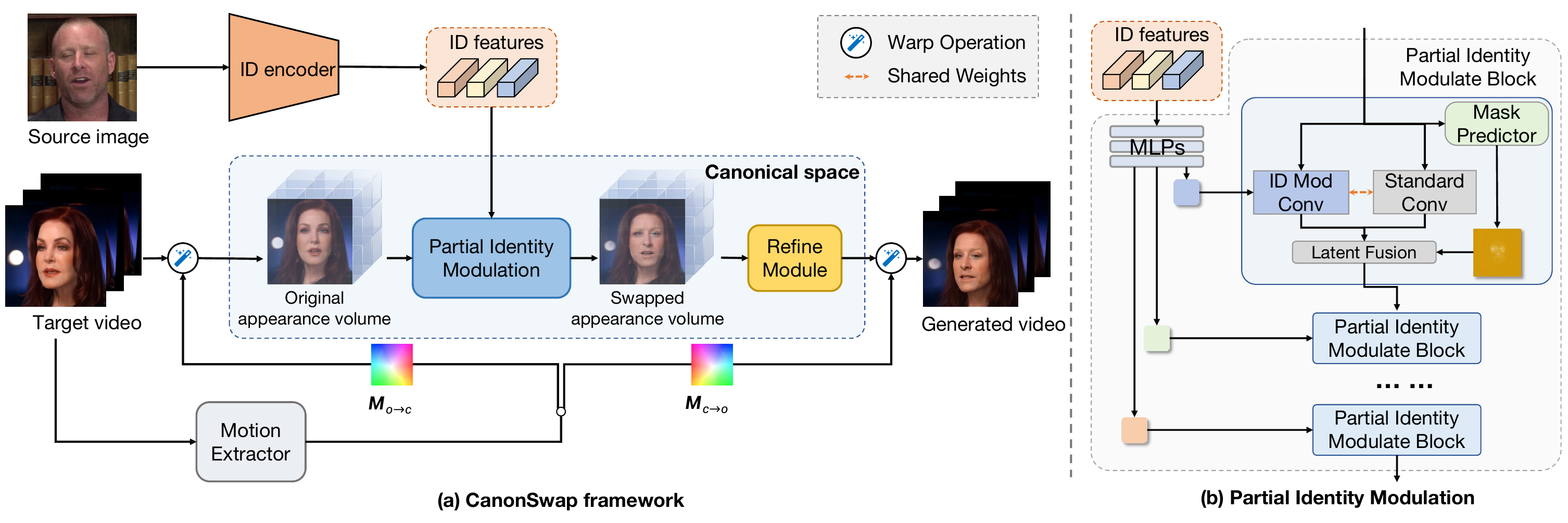}
    \caption{(a) The Pipeline of CanonSwap. Given a source image and a target video, our method first extracts  identity features through an ID encoder from the source image. Each frame of the target video is warped to a canonical space using transformation $\mathbf{M_{o\rightarrow c}}$ estimated by the motion extractor. In this canonical space, we perform identity transfer using the Partial Identity Modulation module. The transformed features are then refined by a refine module. Finally, the refined feature is warped back to the target pose using $\mathbf{M_{c \rightarrow o}}$ and to generate the swapped results. (b) The structure of our Partial Identity Modulation (PIM) module. The PIM module contains several PIM blocks, each block contains two branches, and the outputs of these two branches are fused by a predicted soft spatial mask.}
    \label{fig:pipeline}
\end{figure*}
% Given a source identity image $I_s$ and a target portrait image $I_t$, our goal is to transfer the identity of $I_s$ to $I_t$ while preserving the target's pose and expression. As shown in Figure 2, our framework consists of three main components: (1) an ID encoder that extracts identity features from the source image, (2) a canonical space transformation module that handles pose disentanglement, and (3) a masked identity modulation module that performs identity-preserved feature transformation.

% The overall pipeline operates as follows: First, the source image $I_s$ is processed by an ID encoder to extract multi-scale identity features $\{\mathbf{F_{id}}\}_{i=1}^N$. Simultaneously, the target image $I_t$ undergoes pose removal through our warping module, which estimates the transformation matrix $\mathbf{M_{o→c}}$ to map the target pose to a canonical space. In this canonical space, we apply our masked identity modulation to transfer identity features while preserving structural information. Finally, the transformed features are mapped back to the target pose using the inverse transformation $\mathbf{M_{c→o}}$ and refined through a 3D UNet to generate the final swapped face.

Existing face swapping methods perform face swap directly on the target image or video in its original space. Due to the high coupling between motion and appearance, altering the face often inadvertently modifies the motion, which in turn causes jitters and reduces the overall realism in video face swapping. Therefore, it's necessery to decouple motion information from appearance, which ensures motion consistency while effectively transferring identity information.
% Our key insight is to decouple the motion information from the appearance, thereby reducing the impact of face swapping on the motion. 

The conceptual overview of our approach is illustrated in Fig.~\ref{fig:method}. Given an input video, we first warp it from the original space to a canonical space. In the canonical space, the face retains only appearance information with a fixed and consistent pose. We then perform the face swapping in this canonical space and warp the result back to the original space. Thanks to the decoupling of motion and appearance, CanoSwap can achieve highly consistent and stable swapping results across video frames.
% The conceptual illustration of our approach is shown in Figure~\ref{fig:method}.
% Given an input video, we first warp it from the original space to a canonical space. In the canonical space, the face only contains appearance information and maintains a single, consistent pose. We then perform the face swap in this canonical space before warping the result back to the original space. Although our method processes frames individually, the absence of diverse motion information in the canonical space ensures that the faces remain highly consistent, leading to uniform swapping results. Consequently, we achieve temporally consistent face swapping while precisely preserving the original motion.

% Our pipeline is illustrated in Fig.~\ref{fig:pipeline}(a). It mainly consists of two parts: 
% 1. The Motion Appearance Decoupling module, which employs an implicit 3D keypoint detection combined with warping techniques to accurately remove and add motion while preserving fine details. 
% % To enable two consecutive warping operations (first removing and then adding motion), we propose a refinement module that repairs the features in the canonical space, ensuring precise restoration.
% 2. The face-swapping module in the canonical space. Here, we propose Partial Identity Modulation, a method that precisely and efficiently modifies local features, achieving effective identity transfer while minimizing artifacts.

As shown in Fig.~\ref{fig:pipeline}(a), our method consists of two parts: 1) Canonical Swap Space, which describes how to construct a canonical space for face swapping that eliminates motion information, and how to consistently map the swapped results back to the original space. 2) Partial Identity Modulation, which can accurately and efficiently transform the source identity information into the target appearance features, achieving face swapping in the canonical space. 
%In the following parts, Section~\ref{sec: decouple} introduces the Canonical Swap Space, Section~\ref{sec:partial_modulation} details the Partial Identity Modulation module, and Section~\ref{sec:train} describes our training strategy and loss functions.

% \subsection{Motion Appearance Decoupling}
\subsection{Canonical Swap Space}
\label{sec: decouple}
% Direct face swapping in the original space often causes unintended motion changes due to the tight coupling of appearance and motion. To address this, we first construct a canonical swap space to decouple motion from appearance, and then perform face swapping in this space. The swapped results are subsequently warped back to the original space, preserving dynamic attributes and temporal consistency.
Direct swapping face in the original space usually results in unexpected appearance and motion alterations due to the coupling of appearance and motion. To mitigate this issue, we propose to construct a canonical swap space that decouples motion and appearance, and then conduct face swapping in this space. The swapped results are subsequently warped back to the original space, thereby preserving dynamic attributes and ensuring consistency.
% Due to the strong coupling between appearance and motion, performing face swapping directly in the original space can inadvertently alter the pose. Hence, we propose a motion-appearance decoupling framework: we first warp the video from the original space to a canonical space, carry out face swapping in this canonical space, and finally warp back to the original space.

% This transformation can be realized using portrait animation methods.
%we employ an appearance feature encoder to extract a 4D appearance feature $F_k^o$.
Inspired by~\cite{facev2v},  the canonical swap space can be constructed by using motion-guided warping. We estimate the motion of the target video frame with a motion extractor (please refer to supplementary for details) and obtain the motion transformation $M_{o \rightarrow c}$ and $M_{c \rightarrow o}$.
% Specifically, for the $k$-th frame of the target video in the original space $V_k^o$, we use an implicit keypoint detector to obtain the canonical keypoints $X_k^c \in \mathbb{R}^{n \times 3}$, along with motion deformations, which include pose rotation $X_k^p \in \mathbb{R}^{n \times 3}$, expression $X_k^e \in \mathbb{R}^{n \times 3}$, and translations $X_k^t \in \mathbb{R}^3$, where $n$ denotes the number of keypoints. Using these components, the keypoints for the $k$-th frame are computed as:
% \begin{equation}
% X_k = X_k^c X_k^p + X_k^e + X_k^t.
% \end{equation}
% Then, we feed $X_k$ and $X_k^c$ into a motion estimation module $\mathcal{ME}$ to estimate motion information. By swapping the order of $X_k$ and $X_k^c$, we can simultaneously obtain the deformations from the original space to the canonical space $M_{o \rightarrow c}$, and from the canonical space back to the original space $M_{c \rightarrow o}$:
% \begin{equation}
% M_k^{o \rightarrow c} = \mathcal{ME}(X_k, X_k^c), \quad
% M_k^{c \rightarrow o} = \mathcal{ME}(X_k^c, X_k).
% \end{equation}
Using the estimated motion transformation $M_{o \rightarrow c}$, we can warp the original appearance volume predicted by an appearance encoder to the appearance volume of the canonical space. After face swapping in the canonical space, the swapped appearance volume is then warped back to the original space by $M_{c \rightarrow o}$ and decoded to produce the final result.

% However, traditional animation training typically uses a single driving frame to warp a source image, which creates a mismatch with our pipeline that warps twice in succession. This discrepancy can lead to artifacts when warping back to the original space. 
We note that after two successive warping steps, the appearance volume may contain some discrepancies, leading to artifacts in the final results. Therefore, we propose a refinement module, a lightweight 3D U-Net structure~\cite{unet}, to refine the swapped appearance volume, before warping it back to the original space.

\subsection{Partial Identity Modulation}
\label{sec:partial_modulation}

\noindent
Based on the above mentioned canonical space, we conduct face swapping by modulating the canonical appearance feature.
% Unlike conventional face-swapping methods that inject identity features uniformly across the entire feature map, such indiscriminate modulation can inadvertently alter background and non-facial regions. To overcome this, we propose a \emph{Partial Identity Modulation} framework that selectively applies identity modulation only to facial regions while preserving the rest of the scene.
% Most face-swapping methods rely on injecting identity features at multiple layers of a generator network. However, directly applying identity-related transformations to the entire feature map can introduce unwanted artifacts in background or non-facial regions. To address this challenge, we propose a \emph{Partial Identity Modulation} framework that selectively modulates identity-relevant regions while preserving the rest of the scene.
Many GAN-based methods employ AdaIN for face swapping and achieve promising results. However, AdaIN operates over the entire feature map, often lacks flexibility and can lead to unstable training. Inspired by~\cite{karras2020analyzing}, we further propose a \emph{Partial Identity Modulation} (PIM) module that selectively applies identity modulation only to facial regions while preserving the rest part. By confining the modulation to facial areas, our approach mitigates adversarial effects during training and enhances stability, while the flexible modulation mechanism further enhances the upper bound of face swapping performance. Appendix D proves that our method can achieve faster convergence and effectively mitigate adversarial phenomena in training.

%%%%%%%
As illustrated in Fig.~\ref{fig:pipeline}(b), PIM module contains several blocks, each block contains two parallel branches and adaptively combines the outputs of these branches through a spatial mask $A$, expressed as:
\begin{equation}
    F_{\mathrm{out}} = A \odot F_{\mathrm{id}} + (1 - A) \odot F_{\mathrm{normal}},
\end{equation}
where $\odot$ denotes element-wise multiplication, $F_{\mathrm{id}}$ and $F_{\mathrm{normal}}$ are features generated by two branches. This fusion mechanism enables selective feature modulation across different spatial regions.

% Specifically, we first aggregate multi-scale identity features from ID encoder $F_{\text{id}}^i$ using a series of MLPs to obtain the multi-scale identity code $s_{id}$. The two parallel branches then process the input features as follows:
Specifically, we first aggregate identity features from ID encoder $F_{\text{id}}^i$ using a series of MLPs to obtain the identity code $s_{id}$. The two parallel branches then process the input features as follows:
1) Standard Convolution Branch:
    \begin{equation}
        F_{\text{normal}} = \text{Conv}(F_{in}; \mathbf{W}),
    \end{equation}
    where $\mathbf{W}$ denotes the convolution weights and this branch processes the input feature map $F_{in}$ without any identity-specific transformations.
2) Modulated Convolution Branch:
We first modulate the original convolution weights \(\mathbf{W}\) with the identity code $s_{id}$ and then stabilize the resulting weights via demodulation. This unified process can be formulated as:
\begin{equation}
F_{\mathrm{id}} 
= \mathrm{Conv}\!\Biggl(
F_{in};\,
\frac{s_{id} \cdot \mathbf{W}}{\sqrt{\sum \!\Bigl(s_{id} \cdot \mathbf{W}\Bigr)^{2} + \epsilon}}
\Biggr),
\end{equation}
where \(\epsilon\) is a small constant ensuring numerical stability. In this formulation, the convolution weights are first scaled by \(s_{id}\) to inject identity-specific features, and then normalized by their \(\ell_2\)-norm to prevent excessive variance shifts.
The spatial mask $A \in [0,1]^{H \times W}$ is generated by a mask predictor $\phi(\cdot)$ and can be expressed as:
\begin{equation}
    A = \phi(X).
\end{equation}
This mask ensures that identity modulation primarily affects facial regions (e.g., eyes, nose, and mouth) while preserving the original content in the other irrelevant areas.

Overall, PIM provides fine-grained control over the facial regions, avoiding over-modification and preserving the natural appearance of the target attribute. This selective strategy significantly reduces artifacts in challenging scenarios (e.g., complex backgrounds or large poses), resulting in more realistic and robust results.

\subsection{Training and Loss Functions}
\label{sec:train}
Our warping framework adopts~\cite{facev2v}, and we train the PIM module and refinement in an end-to-end manner. During training, we simultaneously supervise the swapped results in both canonical and original space. The results in the canonical space $I_{s \rightarrow t}^c$ are obtained by decoding the canonical face-swapped features, while the results in the original space $I_{s \rightarrow t}^o$ are obtained by warping the canonical swapped features back and decoding them in the original space.

To ensure accurate identity transfer, we employ identity loss in both canonical and original spaces. The identity loss utilizes a pre-trained face recognition model~\cite{arcface} to measure the identity similarity between the swapped face and the source identity:

\begin{equation}
\begin{aligned}
      \mathcal{L}_{id} = - [\text{Sim}(E_{id}(I_s), E_{id}(I_{s \rightarrow t}^c)) + \\ \text{Sim}(E_{id}(I_s), E_{id}(I_{s \rightarrow t}^o))],
\end{aligned}
\end{equation}
where $E_{id}$ represents a pre-trained face recognition model~\cite{arcface}, $\text{Sim}(\cdot,\cdot)$ denotes the cosine similarity between two feature vectors.

For maintaining structural consistency, we incorporate a perceptual loss that measures the feature-level similarity~\cite{lpips} between the swapped faces and the target faces:
\begin{equation}
    \mathcal{L}_{p} = \mathcal{L}_{LPIPS}(I_{s \rightarrow t}^c, I_t^c) + \mathcal{L}_{LPIPS}(I_{s \rightarrow t}^o, I_t^o),
\end{equation}
where $I_t^c$ represents the target face in canonical space, which can be obtained by decoding canonical feature $F^c_t$.

To preserve pose and expression accuracy, we introduce motion loss, which can be formulated as:
\begin{equation}
    L_{mo} = ||P_{s \rightarrow t} ^c||_1 + ||E_{s \rightarrow t} ^c||_1 + ||P_{s \rightarrow t} ^o - P_t ^o||_1 + ||E_{s \rightarrow t} ^o - E_t ^o||_1,
\end{equation}
where the expression and pose parameters $E$ and $P$ are extracted from the motion extractor.
% The head pose loss ensures consistent pose transfer:
% \begin{equation}
%     \mathcal{L}_{h} = \|M_{c \rightarrow t}^o - M_t\|_1 + \|M_{s \rightarrow t}^c\|_1
% \end{equation}
% while the expression loss maintains natural facial expressions and regularizes the motion information:
% \begin{equation}
%     \mathcal{L}_{e} = \mathcal{L}_{exp}(M_{s \rightarrow t}^c) + \mathcal{L}_{exp\_dis}(M_{s \rightarrow t}^o, M_t)
% \end{equation}

The reconstruction loss, $\mathcal{L}_{r}$, is applied to ensure fidelity when the source and target images belong to the same identity. During training, we randomly sample source-target pairs with a $0.3$ probability of sharing the same identity. The reconstruction loss is formulated as:
\begin{equation}
\mathcal{L}_{r}=
\begin{cases}
\|I_{s \rightarrow t}^c - I_t^c\|_1 +  \|I_{s \rightarrow t}^o - I_t\|_1 & \text{if } id(I_s) = id(I_t),\\
0 & \text{otherwise}.
\end{cases}
\end{equation}
where $id(\cdot)$ denotes the identity of input images.

To enhance the visual quality and realism of the generated images, we employ an adversarial loss:
\begin{equation}
    \mathcal{L}_{g} = \mathcal{L}_{adv}(D(I_{s \rightarrow t}^o))+ \mathcal{L}_{adv}(D(I_{s \rightarrow t}^c)).
\end{equation}
where $D$ denotes the discriminator.

Although the above losses enable effective unsupervised learning of the blending regions, where the network can automatically determine appropriate boundaries for face swapping, we observe that overly sharp transitions may introduce artifacts along these boundaries. Therefore, we introduce additional regularization losses to ensure smooth and accurate blending between the swapped regions:
\begin{equation}
    \mathcal{L}_{m} = \mathcal{L}_{tv}(A) + ||A - A_{GT}||_1
\end{equation}
where $A$ represents the predicted mask and $A_{GT}$ is the ground truth mask when available. The total variation loss $\mathcal{L}_{tv}$ computes the sum of absolute differences between neighboring pixels in both horizontal and vertical directions, encouraging spatial smoothness in the predicted mask.

The overall training objective combines these losses with carefully tuned weights:
\begin{equation}
    \mathcal{L}_{total} = \lambda_{id}\mathcal{L}_{id} + \lambda_{p}\mathcal{L}_{p} + \lambda_{mo}\mathcal{L}_{mo} + \lambda_{r}\mathcal{L}_{r} + \mathcal{L}_{g} + \lambda_{m}\mathcal{L}_{m},
\end{equation}
with $\lambda_{id} = \lambda_r = 10$, $\lambda_{p} = \lambda_{mo} = 5 $, and $\lambda_m = 1$. This comprehensive loss function enables our model to achieve high-quality face swapping results with motion consistency and clean identity transfer.

\section{Metrics of Video Face Swapping}
\label{sec:metrics}

Traditional face swapping evaluations typically rely on metrics such as ID similarity, ID retrieval, expression accuracy, pose accuracy, and FID~\cite{chen2020simswap}. While these metrics have been effective for image-based face swapping, they do not capture the unique challenges of video face swapping, such as temporal consistency and audio-lip synchronization. To address this gap, we propose a set of more fine-grained evaluation metrics specifically designed for video face swapping.

Our approach extends the conventional metrics with additional measurements for the eye and lip regions. For the eyes, in addition to the commonly used gaze estimation, we incorporate the Eye Aspect Ratio (EAR)~\cite{EAR} to more accurately assess blink patterns. 
% and subtle eye movements. 
For the lip region, we introduce synchronization (sync) metrics~\cite{syncnet}. Specifically, we adopt Lip Sync Error-Distance (LSE-D) and Lip Sync Error-Confidence (LSE-C) from the talking head synthesis task to evaluate how well the lip movements align with the audio. LSE-D quantifies the average deviation of lip landmarks from the ground truth, while LSE-C measures the confidence of the lip synchronization predictions.

To support these comprehensive evaluations, we also introduce a new benchmark named VFS (Video Face Swapping benchmark). The VFS benchmark comprises 100 source-target pairs randomly sampled from the VFHQ dataset~\cite{xie2022vfhq}. Each target video includes the first 100 frames along with 4 seconds of corresponding audio, allowing for a thorough assessment of both visual fidelity and audio-lip synchronization.

\section{Experiment}
\label{experiment}
\subsection{Experimental Settings}
\subsubsection{Datasets}
We train our model on the VGGFace dataset~\cite{cao2018vggface2}, a widely-used face recognition dataset. We perform face detection on the original VGGFace dataset and filter out images with width less than 130 pixels, resulting in 930K images. For training, these images are resized to  $512 \times 512$ resolution.
% To evaluate model performance, we initially considered the commonly used FF++ dataset. However, we found it inadequate for assessing video face swapping quality, particularly in real-world applications where audio-lip synchronization plays a crucial role, which is often overlooked in both image-based and current video-based face swapping works. Therefore, we introduce a new benchmark named VFS specifically designed for video face swapping evaluation. This benchmark consists of 100 source-target pairs randomly sampled from VFHQ dataset, where each target video contains the first 100 frames along with 4 seconds of corresponding audio, enabling comprehensive assessment of both visual quality and audio-lip synchronization.
We evaluate our model's performance on two datasets: the widely-used FaceForensics++ (FF++) dataset~\cite{rossler2019faceforensics++} and our newly proposed VFS benchmark. 
% FF++ provides a solid basis for comparing visual quality, while VFS, consisting of 100 source-target pairs from the VFHQ dataset with 100 target frames and 4 seconds of audio per video, enables a comprehensive evaluation of both visual fidelity and audio-lip synchronization.

\subsubsection{Compare Methods}
To demonstrate the effectiveness of our method, we compare our method with GAN-based methods like SimSwap~\cite{chen2020simswap}, FSGAN~\cite{nirkin2022fsganv2} and E4S~\cite{liu2023fine}, and Diffusion-based methods like DiffSwap~\cite{zhao2023diffswap}, REFace~\cite{REFace} and Face Adapter~\cite{faceadapter}.

\subsection{Quantitative Evaluations}
\subsubsection{Overall Metrics}
We evaluate our method using two distinct test sets: FF++ dataset and our VFS benchmark. On FF++, we follow conventional face swapping evaluation protocols with five established metrics: ID retrieval, ID similarity, pose accuracy, expression accuracy, and Fréchet Inception Distance (FID). ID retrieval and similarity are computed using a face recognition model~\cite{Shiohara_2023_ICCV} with cosine similarity.
% , where we match against a single frame from each video to increase evaluation stringency. 
Pose accuracy~\cite{liu2025teaser} is measured by the Euclidean distance between the estimated and ground truth poses, while expression accuracy~\cite{liu2025teaser} is computed as the L2 distance between the corresponding expression embeddings.

For our video benchmark, we 
% introduce more fine-grained metrics tailored to video face swapping quality assessment. We 
employ Fréchet Video Distance (FVD) instead of FID to better evaluate temporal consistency. We implement motion jitter analysis (Temporal Consistency, a.k.a TC) by comparing optical flow fields between source and swapped videos to quantify unnatural facial movements. We also adopt fine-grained metrics in Section~\ref{sec:metrics}: Gaze and EAR computed from facial landmarks, along with LSE-D and LSE-C measured using SyncNet~\cite{syncnet}.
% Furthermore, recognizing that eye movements and mouth dynamics are crucial for natural face swapping, we introduce specialized metrics for these aspects. For eye region evaluation, we employ gaze direction estimation and Eye Aspect Ratio (EAR) to measure the accuracy of eye movements and blinking patterns. For mouth movement assessment, we incorporate Lip Sync Error-Distance (LSE-D) and Lip Sync Error-Confidence (LSE-C) from the talking head synthesis domain. LSE-D measures the average distance between predicted and ground truth lip landmarks synchronized with audio, while LSE-C evaluates the confidence of lip sync prediction. These lip sync metrics are particularly important as our empirical observations indicate that poor audio-lip synchronization is a major factor contributing to unrealistic face swapping results in practical applications.
\begin{table}
\centering
\begin{tabular}{@{}l@{\hspace{6pt}}c@{\hspace{6pt}}c@{\hspace{6pt}}c@{\hspace{6pt}}c@{\hspace{6pt}}c@{}}
\toprule
Method & ID Sim.$\uparrow$ & ID R.$\uparrow$ & Pose$\downarrow$ & Exp$\downarrow$ & FID$\downarrow$ \\
\midrule
SimSwap~\cite{chen2020simswap} & \underline{0.5416} & \underline{97.91} & 0.0158 & 0.9658 & \underline{7.44} \\
FSGAN~\cite{nirkin2022fsganv2} & 0.2781 & 41.35 & 0.0156 & \textbf{0.7184} & 14.58 \\
DiffSwap~\cite{zhao2023diffswap} & 0.3179 & 48.93 & \underline{0.0142} & 0.7370 & 10.80 \\
E4S~\cite{liu2023fine} & 0.4435 & 86.67 & 0.0212 & 1.0751 & 24.66 \\
Face Adapter~\cite{faceadapter} & 0.5035 & 94.46 & 0.0197 & 1.0064 & 16.01 \\
REFace~\cite{REFace} & 0.4632 & 91.37 & 0.0201 & 1.1612 & 18.56 \\ 
\midrule
CanonSwap & \textbf{0.5751} & \textbf{98.29} & \textbf{0.0119} & \underline{0.7328} & \textbf{6.21} \\
\bottomrule
\end{tabular}
\caption{Quantitative comparison on FF++. Our approach demonstrates superior performance on virtually all metrics, while maintaining competitive results in expression metric.}
\label{tab:comparisonff}
\end{table}

\begin{figure*}
    \centering
    \includegraphics[width= \linewidth]{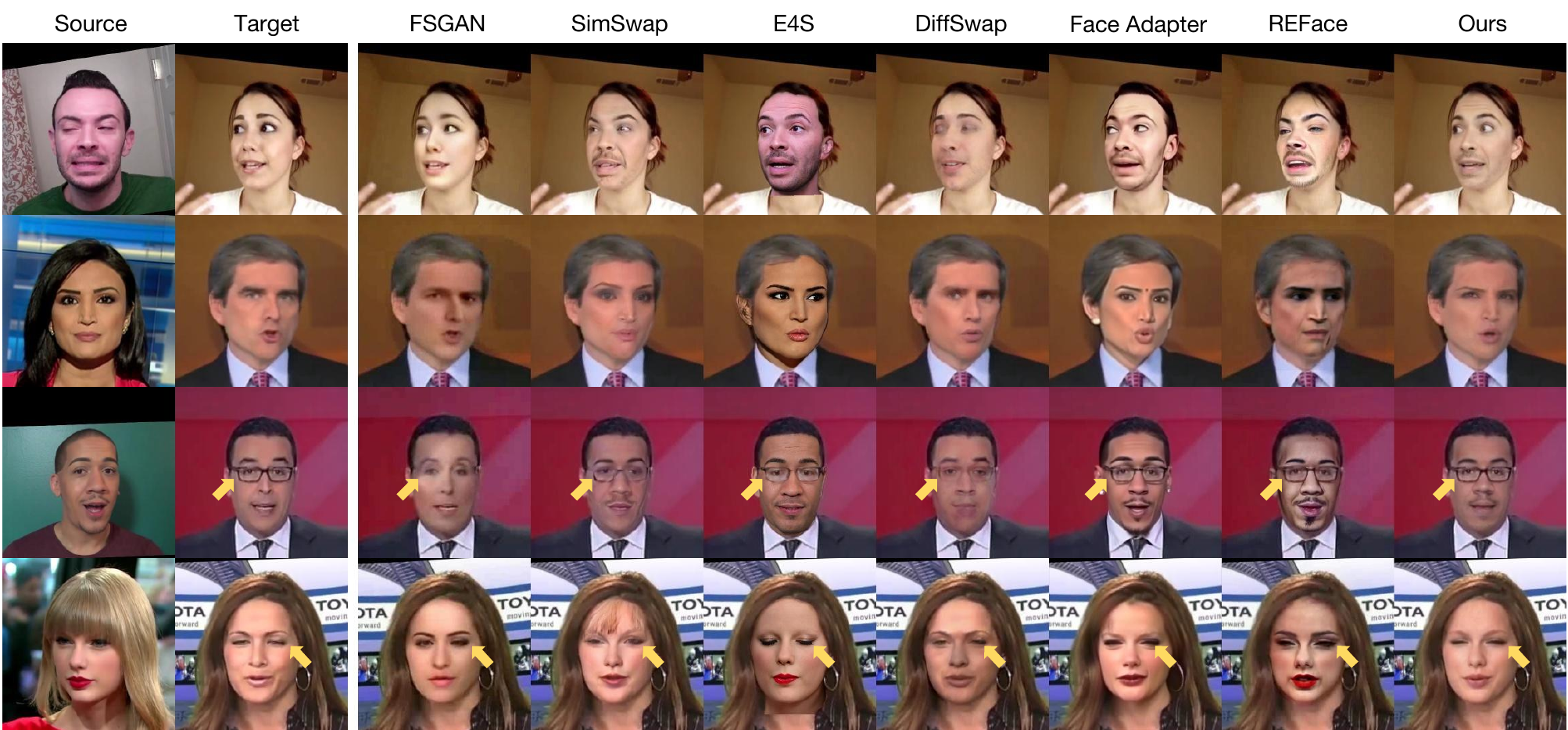}
    \caption{Qualitative results on FF++. Our method achieves accurate identity transfer while ensuring precise motion consistency, without introducing visible artifacts. See enlarged views for details.}
    \label{fig:qualitativeff}
\end{figure*}
\subsection{Qualitative Evaluations}

\begin{table*}
\centering
\begin{tabular}{l|cccc|cc|cc|cc}
\toprule
\multirow{2}{*}{Method} & \multicolumn{4}{c|}{Global} & \multicolumn{2}{c|}{Eyes} & \multicolumn{2}{c|}{Mouth(Sync)} & \multicolumn{2}{c}{Quality} \\
\cline{2-11}
 & ID sim.$\uparrow$ & ID R.$\uparrow$ & Pose$\downarrow$ & Exp.$\downarrow$ & Gaze$\downarrow$ & EAR$\downarrow$ & LSE-D$\downarrow$ & LSE-C$\uparrow$ & TC$\downarrow$ & FVD$\downarrow$ \\
\hline
SimSwap~\cite{chen2020simswap} & 0.5160 & \underline{98.87} & \underline{0.0196} & 0.9495 & 0.1206 & 5.403 & \underline{8.344} & \underline{5.306} & 0.773 & \underline{136.78} \\
FSGAN~\cite{nirkin2022fsganv2} & 0.1442 & 29.30 & 0.0204 & \textbf{0.7525} & \underline{0.1037} & \underline{3.976} & 8.847 & 4.710 & \underline{0.760} & 322.30 \\
DiffSwap~\cite{zhao2023diffswap} & 0.2461 & 63.45 & 0.0281 & 0.8646 & 0.1187 & 4.535 & 10.670 & 3.213 & 0.959 & 508.16 \\
E4S~\cite{liu2023fine} & 0.3953 & 89.63 & 0.0288 & 1.1780 & 0.1423 & 6.150 & 9.554 & 3.913 & 1.066 & 377.48 \\
Face Adapter~\cite{faceadapter} & \underline{0.5215} & 98.77 & 0.0229 & 1.0354 & 0.1190 & 6.270 & 9.309 & 4.399 & 1.312 & 424.61 \\
REFace~\cite{REFace} & 0.4306 & 96.23 & 0.0245 & 1.1782 & 0.1514 & 7.148 & 10.281 & 3.214 & 1.268 & 400.88 \\ \hline
CanonSwap & \textbf{0.5748} & \textbf{99.78} & \textbf{0.0159} & \underline{0.7592} & \textbf{0.0928} & \textbf{3.742} & \textbf{7.938} & \textbf{6.053} & \textbf{0.513} & \textbf{125.30} \\
\bottomrule
\end{tabular}
\caption{Quantitative comparison on our VFS benchmark. We achieve the best performance across all metrics, and our expression results are nearly on par with the top-performing method.}
\label{tab:comparisonvfs}
\end{table*}

\subsubsection{Evaluation Results}
The evaluation results on both VFS benchmark and FF++ dataset are presented in Tab.~\ref{tab:comparisonff} and Tab.~\ref{tab:comparisonvfs}. The quantitative results demonstrate that our method consistently outperforms existing GAN-based methods and diffusion-based approaches across multiple metrics.
In terms of identity preservation, our method achieves the highest ID similarity score and ID retrieval accuracy on both datasets, showing significant improvements over both kinds of approaches. 
In addition, our method yields the lowest errors on pose metrics and competitive results on expression metrics, demonstrating the most precise motion alignment with the target video's motion compared to existing methods.

We also have a significant improvement in mouth synchronization metrics, where our method achieves $\mathbf{7.938}$ and $\mathbf{6.053}$ on LSE-D and LSE-C respectively, surpassing both GAN-based and diffusion-based methods. This superior lip synchronization ability is also reflected in the quality metrics, where our method achieves the best FID and FVD scores, significantly outperforming other methods, demonstrating better synthesis quality while maintaining temporal consistency. These results show that decoupling appearance and motion is essential for video face swapping.

\begin{figure*}
    \centering
    \includegraphics[width=\linewidth]{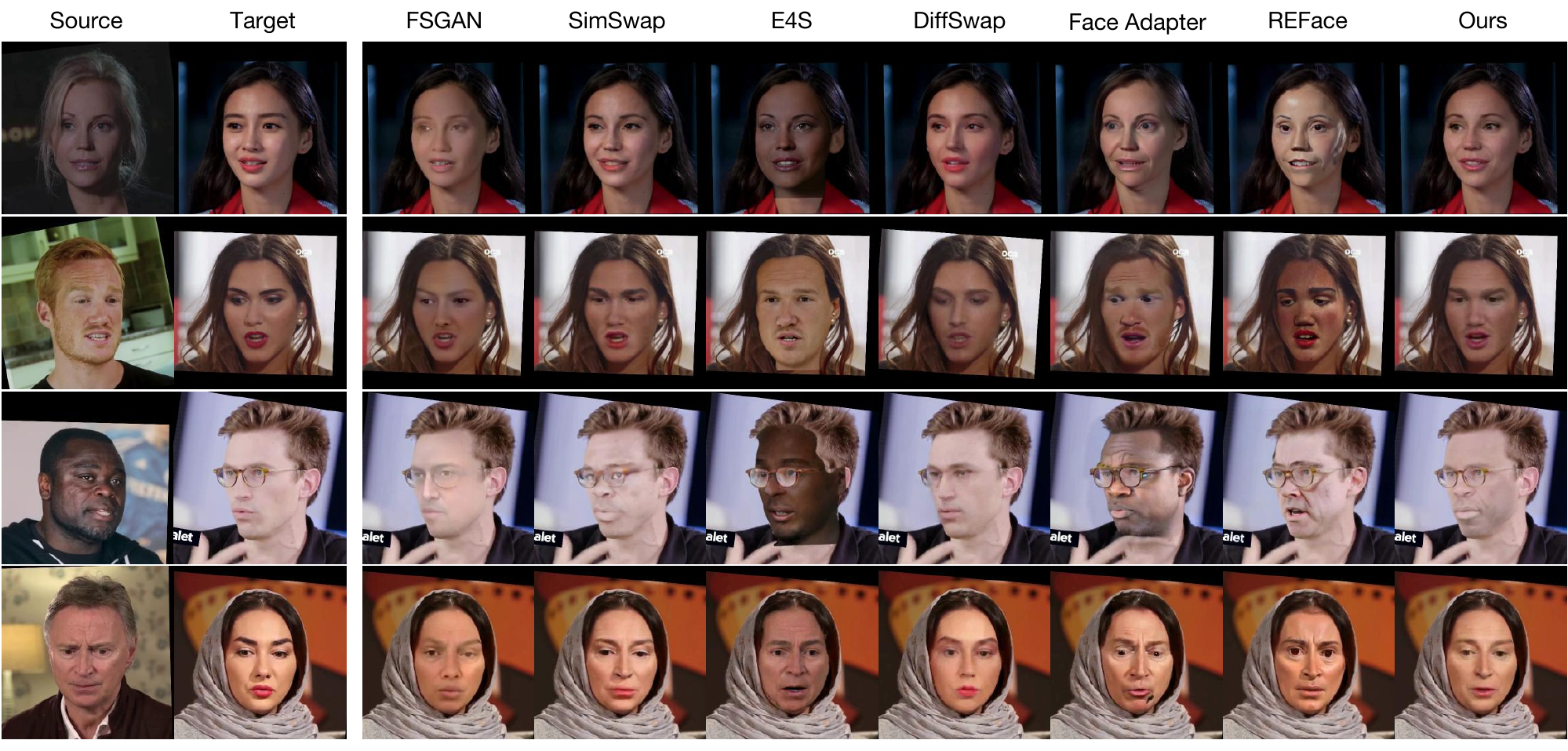}
    \caption{Quantitative results on the VFS benchmark. Our method achieves accurate identity transfer while ensuring precise motion consistency, without introducing visible artifacts. See enlarged views for details.}
    \vspace{-0.5cm}
    \label{fig:qualitativevfs}
\end{figure*}

% Figures~\ref{fig:qualitativeff} and~\ref{fig:qualitativevfs} showcase the qualitative results of our method compared to several state-of-the-art face-swapping approaches, including FSGAN, SimSwap, E4S, DiffSwap, Face Adapter, and REFace. 
To further evaluate the effectiveness of CanonSwap, we conduct qualitative comparisons on the FF++ and VFS benchmarks, as shown in Fig.~\ref{fig:qualitativeff} and Fig.~\ref{fig:qualitativevfs}. The results show that our method not only achieves accurate identity transfer but also preserves precise motion alignment.

% \subsubsection{User Study}
\begin{figure}
    \centering
    \includegraphics[width=\linewidth]{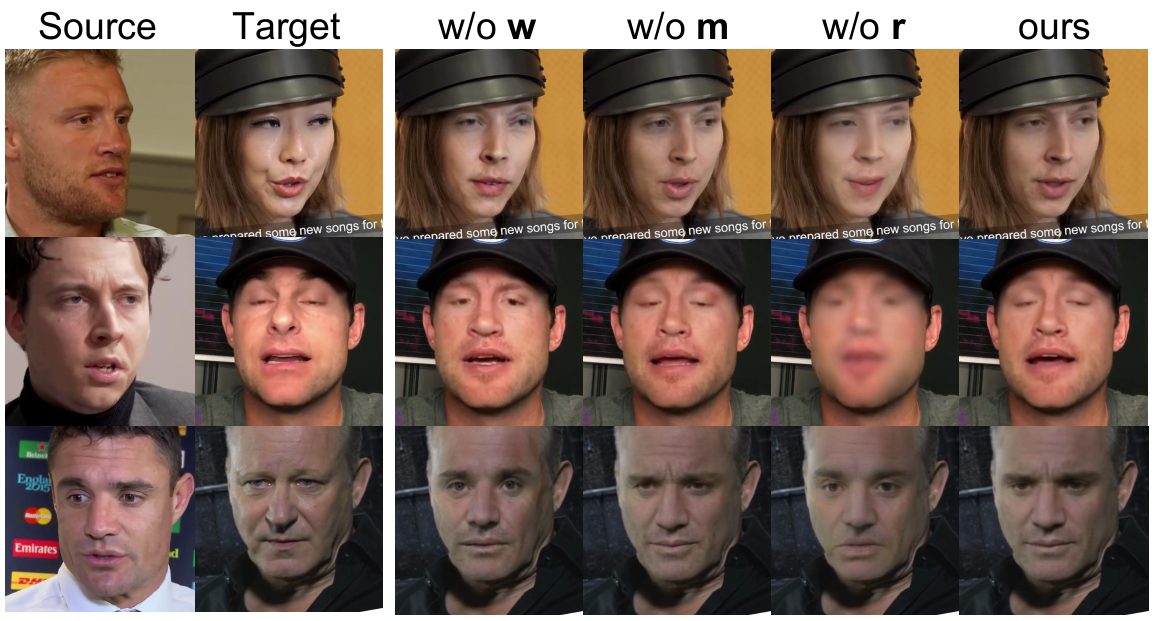}
    \caption{Qualitative results of ablation study.  We compare the results of removing each module individually: (1) omitting the warping step (w/o \textbf{w}), (2) removing the spatial mask (w/o \textbf{m}), and (3) excluding the refinement module (w/o \textbf{r}). 
    % Our complete method exhibits the best overall visual quality, accurately transferring identity while preserving pose alignment and minimizing artifacts.
    }
    \label{fig:ablation}
\end{figure}
\subsection{Ablation Study}

% We perform an ablation study on the VFS benchmark to investigate the importance of each component in our pipeline (Table~\ref{tab:comparisonff} and Figure~\ref{fig:ablation}). Specifically, we remove three key modules individually: (1) \textbf{w/o w} removes the warping step and thus does not decouple motion and appearance, (2) \textbf{w/o m} removes the soft spatial mask and applies modulation globally, and (3) \textbf{w/o r} omits the refinement module that repairs canonical-space features before warping back to the original space. As shown in Table~\ref{tab:comparisonff}, removing any of these components degrades performance across all metrics. In particular, \textbf{w/o r} suffers from a drastic drop in identity similarity (ID Sim.) and a significant increase in FVD, underscoring the importance of refining the canonical-space features to maintain stable face swapping. Meanwhile, \textbf{w/o m} leads to noticeable artifacts, as applying modulation globally introduces unwanted changes to identity-irrelative regions, and \textbf{w/o w} shows weaker pose and expression alignment due to the lack of motion-appearance disentanglement. 

We conduct ablation study on the VFS benchmark to evaluate the impact of each module in our pipeline (see Tab.~\ref{tab:ablation} and Fig.~\ref{fig:ablation}). Specifically, we remove three components individually: (1) w/o \textbf{w} omits the warping step, directly conducting face swarping in the original space; (2) w/o \textbf{m} removes the soft spatial mask, resulting in global modulation across the entire feature map; and (3) w/o \textbf{r} excludes the refinement module that enhances canonical-space features before warping back. As shown in Tab.~\ref{tab:ablation}, removing any of these components degrades the performance across multiple metrics. Fig.~\ref{fig:ablation} further illustrates these issues qualitatively: w/o \textbf{w} fails to align pose and expression accurately, w/o \textbf{m} introduces more undesired textures, and w/o \textbf{r} leads to blurry or inconsistent identity details. These results demonstrate that all three modules are essential for achieving accurate identity transfer, precise pose alignment, and artifact-free results in video face swapping.

\begin{table}
\centering
\begin{tabular}{@{}l@{\hspace{6pt}}c@{\hspace{6pt}}c@{\hspace{6pt}}c@{\hspace{6pt}}c@{\hspace{6pt}}c@{}}
\toprule
Method & ID Sim.$\uparrow$ & ID R.$\uparrow$ & Pose$\downarrow$ & Exp$\downarrow$ & FVD$\downarrow$ \\
\midrule
w/o \textbf{w} & 0.5702 & 99.41 & 0.0227 & 0.9512 & 131.57 \\
w/o \textbf{m} &    0.5508    &   97.38    &   0.0162     &    0.7669    &    165.17   \\
w/o \textbf{r} &    0.4778    &   94.73    &     0.0264   &     1.0241    &    481.77    \\
\midrule
Ours & \textbf{0.5748} & \textbf{99.78} & \textbf{0.0159} & \textbf{0.7592} & \textbf{125.30} \\
\bottomrule
\end{tabular}
\caption{Quantitative results of ablation study on the VFS benchmark. Each row omits a component in our pipeline: warping (\textbf{w}), masking (\textbf{m}), and refinement (\textbf{r}). The proposed three components are essential for high-quality video face swapping.}
\vspace{-0.5cm}
\label{tab:ablation}
\end{table}

\subsection{Face Swapping and Animation}
\begin{figure}
    \centering
    \includegraphics[width=\linewidth]{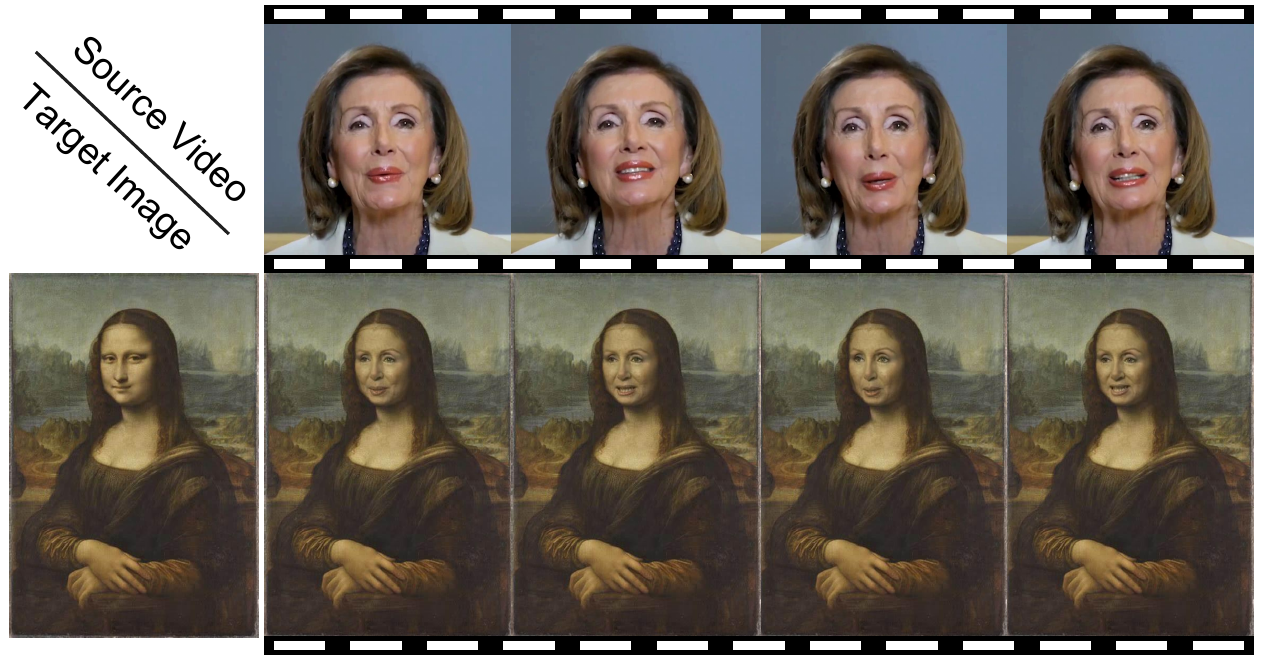}
    \caption{Face swapping and animation results. Both identity and expression of result video come from the source video.}
    \vspace{-0.2cm}
    \label{fig:video2image}
\end{figure}

In our CanonSwap, the input image is decoupled into two components: appearance and motion (i.e. pose and expression). 
% The motion is guaranteed by warping the appearance feature using the extracted pose and expression. 
Thus, other than changing the appearance, CanonSwap also supports altering expressions and poses. Specifically, during the warping-back process, the expression of the target can be replaced with that of the source, allowing for simultaneous identity and expression transfer. This capability enables both face swapping and facial animation within a single framework. As shown in Fig.~\ref{fig:video2image}, CanonSwap not only performs face swapping, but also animates the target image, making it mimic the expressions and actions of the source, thus broadening its potential applications.

\section{Conclusion}
\label{conclusion}

% In this work, we propose a novel video face swapping framework that fundamentally resolves temporal instability issues by decoupling pose variations from identity transfer. The key innovation lies in performing face swapping in a canonical space where facial poses are decoupled, eliminating motion-induced artifacts that plague traditional approaches. Our partial identity modulation module enables precise control over the swapping region, ensuring high-quality results while maintaining temporal consistency when warping back to original poses. To facilitate comprehensive evaluation, we introduce fine-grained synchronization metrics. Extensive experiments demonstrate that our method significantly advances the state-of-the-art in video face swapping, particularly in achieving stable and realistic results across varying poses and expressions.

We propose a novel video face swapping framework that resolves temporal instability by decoupling pose variations from identity transfer in canonical space. Our partial identity modulation module enables precise swapping control while maintaining temporal consistency. We introduce fine-grained synchronization metrics for evaluation. Extensive experiments demonstrate significant advances in stable and realistic video face swapping across varying poses and expressions.

{
    \small
    \bibliographystyle{ieeenat_fullname}
    \bibliography{main}
}
\clearpage
\appendix
\maketitlesupplementary

\section{Training Details}
Our method is implemented in PyTorch and trained on two NVIDIA A6000 GPUs, with a batch size of 6 per GPU. We use the AdamW optimizer (weight decay = \(1\times10^{-4}\), \(\beta_1=0.5\), \(\beta_2=0.999\)) for both generator and discriminator and set the initial learning rate to \(1\times10^{-4}\). The model is trained for 150k steps in  total.

For discriminator, we adopt the same architecture as SPADE. During training, we introduce an additional gradient penalty loss, which enforces smooth decision boundaries by penalizing large gradients in the discriminator. This penalty stabilizes training and helps the discriminator better distinguish between real and generated samples.

\section{Visualization of Canonical Space}
\begin{figure}[H]
    \centering
    \includegraphics[width=\linewidth]{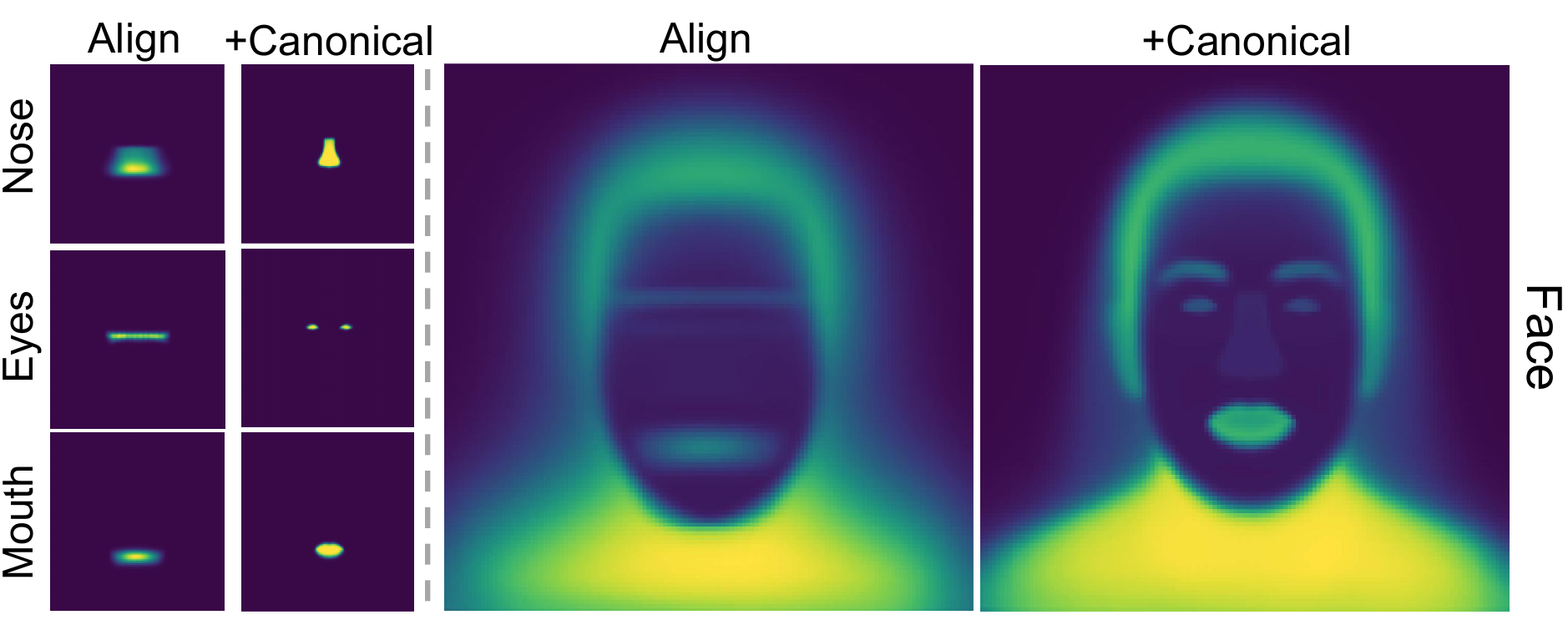}
    \caption{Comparison between traditional face alignment (left) and our canonical-space transformation (right), visualized by averaging segmentation maps across multiple samples. In traditional alignment, residual motion information causes blurred and inconsistent boundaries. By contrast, our canonical-space transformation effectively decouples motion, resulting in more uniform and clearly defined facial regions.}
    \label{fig:canonical}
\end{figure}

To provide an intuitive illustration of how our canonical space appears after motion decoupling, we randomly select 10k frames from our CVF benchmark and apply a crop-and-align procedure to obtain \emph{Align Set}. Next, we transform the images in \emph{Align Set} into the canonical space, yielding \emph{+Canonical Set}. We then use a face segmentation model to compute the average parsing map for each set, as well as individual nose, eyes, and mouth regions, and visualize the results in Fig.~\ref{fig:canonical}. As shown, the canonical space removes motion information, causing facial features to align almost perfectly. By contrast, the standard alignment method still contains motion, resulting in blurred parsing boundaries—particularly around the eyes, which can shift over a wide range.

\begin{figure}
    \centering
    \includegraphics[width= \linewidth]{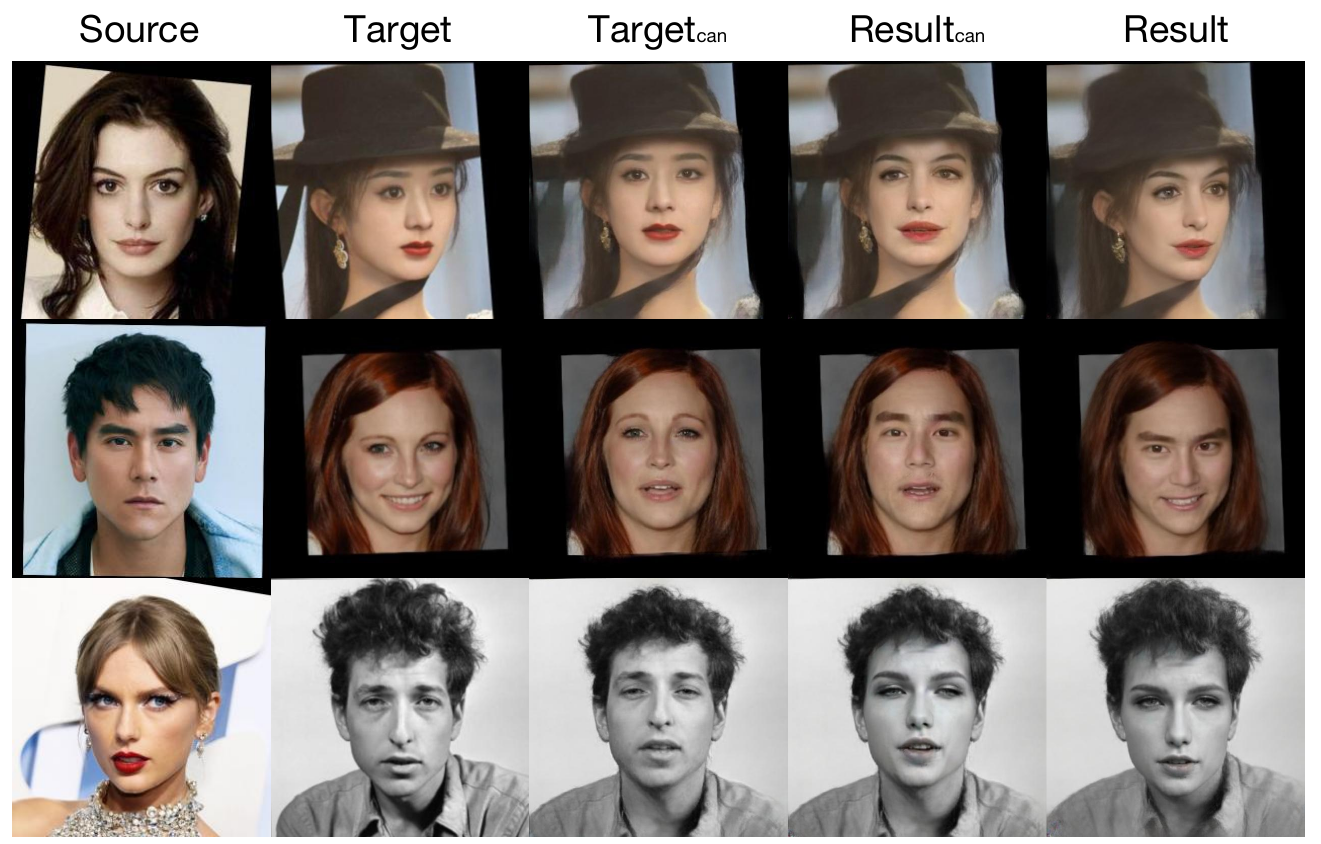}
    \caption{Visualization of outputs of each stage of CanonSwap.}
    \label{fig:midout}
\end{figure}

Furthermore, we visualize the outputs of each stage of CanonSwap, as shown in Fig.~\ref{fig:midout}.

\section{Details of Motion Extractor}
\begin{figure}
    \centering
    \includegraphics[width=\linewidth]{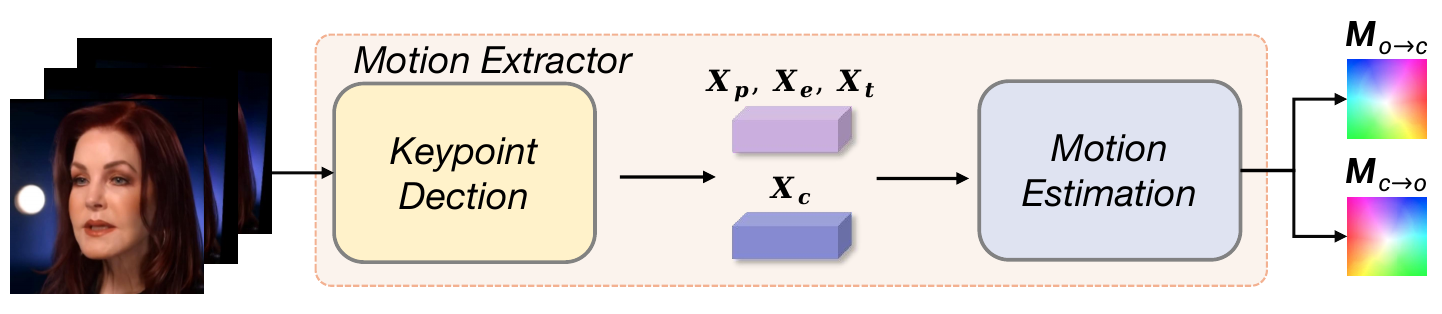}
    \caption{The components of our motion extractor.}
    \label{fig:me}
\end{figure}
The details of motion extractor is shown in Fig.~\ref{fig:me}, 
specifically, for a frame of the target video in the original space $V_o$, we use an implicit keypoint detector to obtain the canonical keypoints $X_c \in \mathbb{R}^{n \times 3}$, along with motion deformations, which include pose rotation $X_p \in \mathbb{R}^{n \times 3}$, expression $X_e \in \mathbb{R}^{n \times 3}$, and translations $X_t \in \mathbb{R}^3$, where $n$ denotes the number of keypoints. Using these components, the keypoints for the frame are computed as:
\begin{equation}
X = X_c X_p + X_e + X_t.
\end{equation}
Then, we feed $X$ and $X_c$ into a motion estimation module $\mathcal{E}$ to estimate motion information. By swapping the order of $X$ and $X_c$, we can simultaneously obtain the deformations from the original space to the canonical space $M_{o \rightarrow c}$, and from the canonical space back to the original space $M_{c \rightarrow o}$:
\begin{equation}
\label{eq:hh}
M_{o \rightarrow c} = \mathcal{E}(X, X_c), \quad
M_{c \rightarrow o} = \mathcal{E}(X_c, X).
\end{equation}

\begin{figure*}
    \centering
    \includegraphics[width= \linewidth]{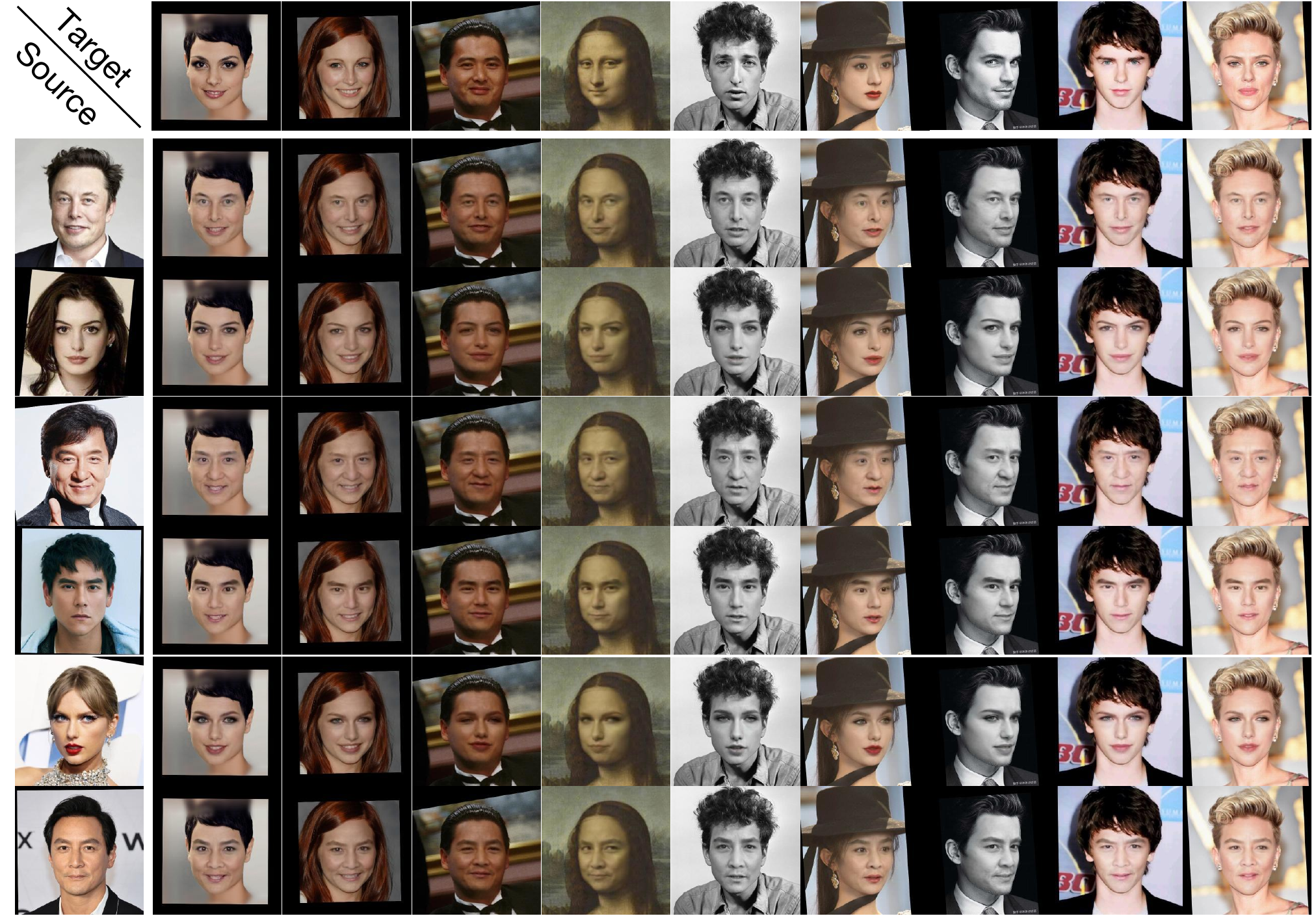}
    \caption{more qualitative results through a face matrix.}
    \label{fig:matrix}
\end{figure*}

\section{Advantages of the PIM.}
\begin{figure}
    \centering
    \includegraphics[width=\linewidth]{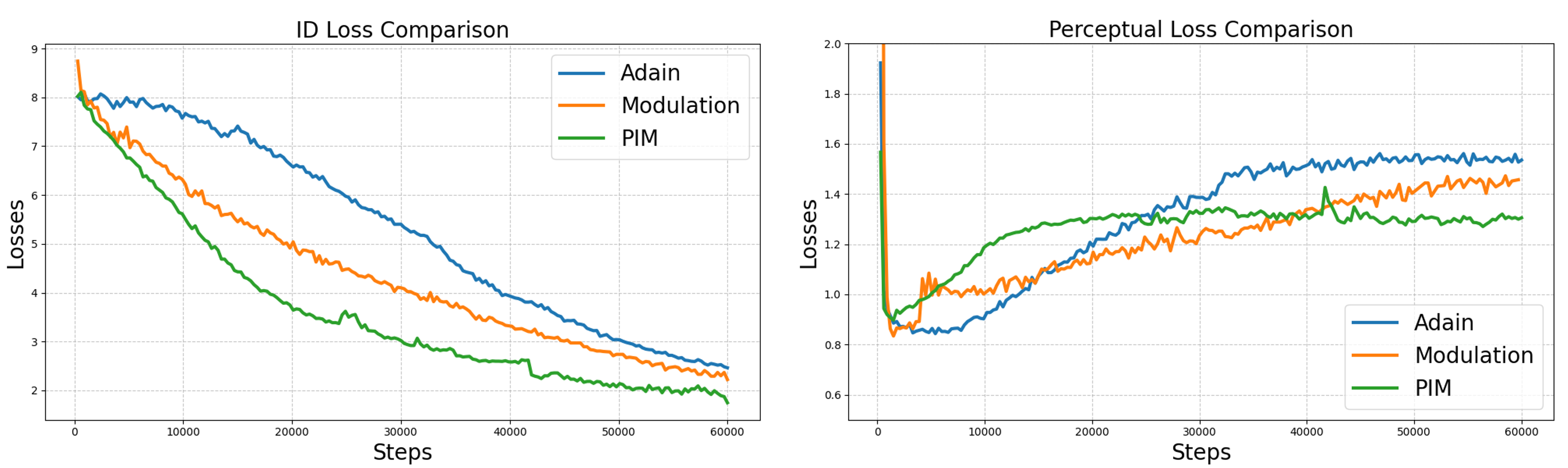}
    \caption{Training loss curves on the same dataset, our PIM achieves the fastest convergence rate and demonstrates lower ID Loss and Perceptual Loss, effectively mitigating the adversarial relationship between losses and achieving a higher performance ceiling.}
    \label{fig:trainloss}
\end{figure}

Our PIM module addresses a key drawback of traditional AdaIN/modulation-based methods—their global application alters identity-irrelevant regions, which is suboptimal for face swapping. This often leads to (1) \textbf{visible artifacts} and (2) \textbf{unstable training from conflicting losses}, the latter often overlooked. We compare AdaIN, global modulation, and PIM under the same setting. As shown in Fig.~\ref{fig:trainloss}, PIM converges faster and alleviates the conflict between identity loss and perceptual loss (lowest ID loss and lowest perceptual loss), resulting in better overall performance and a higher optimization ceiling.

\section{Computational Efficiency}
We evaluate inference efficiency by comparing our method with existing approaches, as shown in the table below (FPS). Our methods is faster than \Diffuison{Diffusion}/\StyleGan{StyleGAN}-based methods.
\begin{table}[htbp]
  \centering
  \label{tab:face_swap_comparison_rotated}
  \footnotesize
  \setlength{\tabcolsep}{0.8pt}
  \renewcommand{\arraystretch}{0.7}

  \begin{tabular}{cccccccc}
    \toprule
     % & \multicolumn{2}{c}{\scriptsize GAN} & {\scriptsize StyleGAN} & \multicolumn{3}{c}{\scriptsize Diffusion} & \\[-2pt]  % 行高减小
    % \midrule
    Metrics & \Gan{Simswap} & \Gan{FSGAN} & \StyleGan{E4S} & \Diffuison{Diffswap} & \Diffuison{FaceAdapter} & \Diffuison{REFace} & Ours\\
    \midrule
    FPS & 16 & 21 & 4 & 0.11  & 0.35 & 0.21 & 14 \\
    \bottomrule
  \end{tabular}
\end{table}

\section{Face Swapping and Animation}
% In the main text, we briefly explain how to swap a video with an image by transferring the source video's expression to the target image. 
To achieve face swapping and animation, we need to change the warping back deformation $M_{c \rightarrow o}$ in Eq.~\ref{eq:hh}.
Specifically, we obtain \(X_c\), \(X_p\), \(X_e\), and \(X_t\) from the target frame, and also extract the source’s expression from the source frame. During the transformation from the original space to the canonical space, we follow the procedure described in the main text. In the warp-back stage, we compute a new keypoint \(X'\) as
\begin{equation}
X' = X_p X_c + X_e^s + X_t,
\end{equation}
where \(X_e^s\) denotes the source’s expression. We then use the motion estimator to obtain a new warp deformation,
\begin{equation}
M_{c \rightarrow o}' = \mathcal{E}\bigl(X_c, X_2\bigr),
\end{equation}
and apply it to warp back, thereby transferring the source expression to the target image. 

\section{More Qualitative Results}
\begin{figure}
    \centering
    \includegraphics[width=\linewidth]{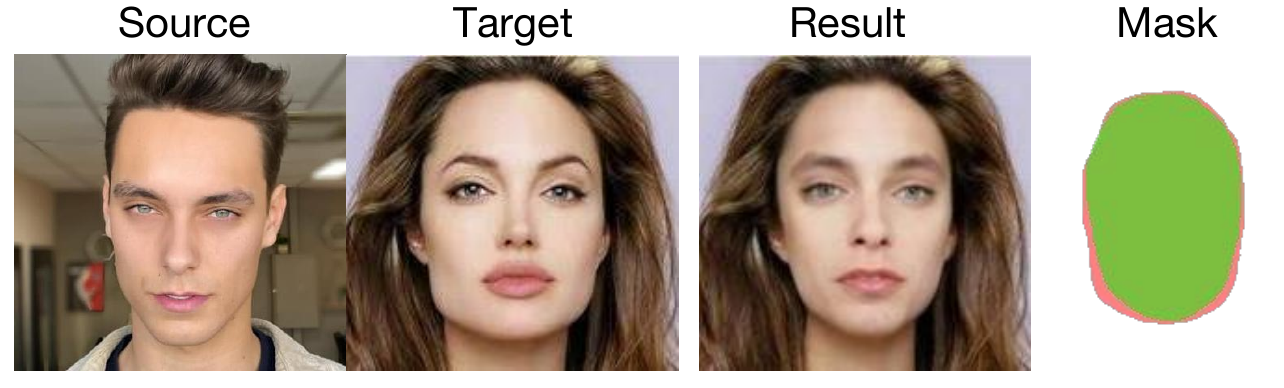}
    \caption{By exchanging canonical keypoints, our method can also achieve shape transfer to some extent.}
    \label{fig:shape}
\end{figure}

\begin{figure}
    \centering
    \includegraphics[width=\linewidth]{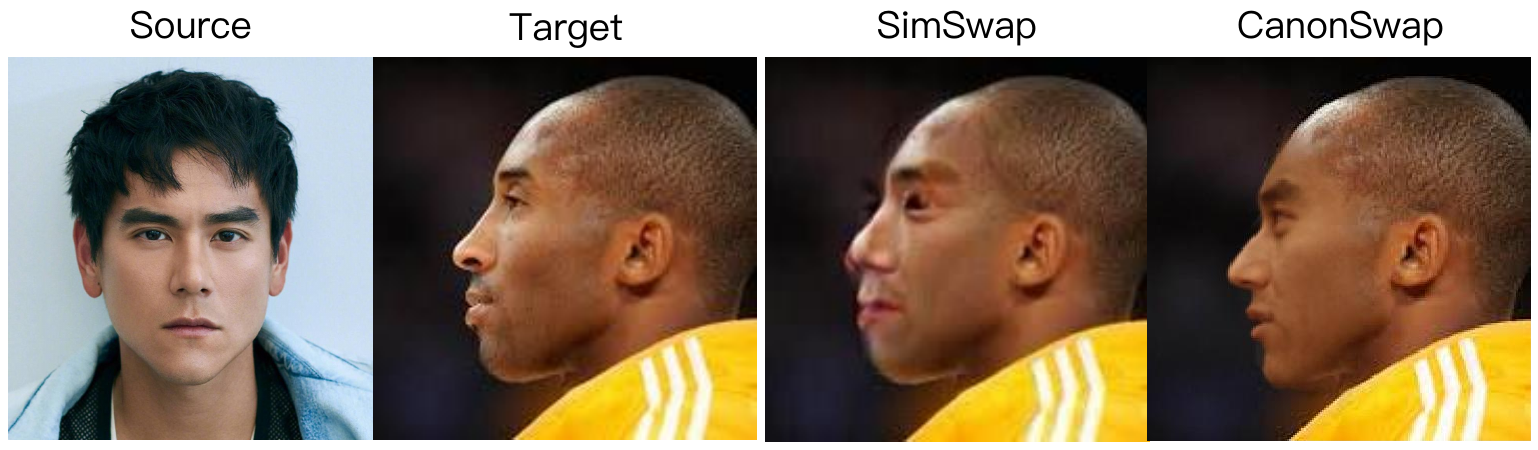}
    \caption{Qualitative comparison in large pose variation situation.}
    \label{fig:large_pose}
\end{figure}

To demonstrate the robustness of our model, we conducted a matrix swap, and the results are shown in Fig.~\ref{fig:matrix}. Furthermore, compared to existing face swapping methods, our approach can leverage powerful animation priors to maintain robust performance under large pose variations. Moreover, by replacing the target's canonical keypoints with those of the source, the facial geometry can be adaptively aligned to match the source's structure to some extent, which is shown in Fig.~\ref{fig:shape}. We also conduct an evaluation in large pose variation situation, which is shown in Fig.~\ref{fig:large_pose}. Warping-based animation (e.g., talking head) may struggle with extreme pose variations due to insufficient target-pose features. In contrast, CanonSwap performs face swapping in a canonical pose and warps back to the original pose while preserving the original pose features. This enables robust performance under large pose variation. As shown in Fig. (a), CanonSwap outperforms prior methods in such scenarios, where SimSwap typically fails to handle large pose differences.

\section{Ethical Considerations}
This research is conducted solely for academic purposes and to advance the video face swapping technology. We use publicly available datasets and adhere to ethical guidelines in our experimentation. While our work aims to improve the fidelity and temporal consistency of face swapping, we acknowledge the potential for misuse in applications such as deepfakes and identity manipulation. We strongly advocate for responsible use of this technology and caution against applications that may infringe on privacy, consent, or intellectual property rights. Researchers and practitioners are encouraged to consider the ethical implications and to implement safeguards to prevent harmful or deceptive uses of our methods.

\end{document}